\documentclass[runningheads,a4paper]{llncs}
\makeatletter
\@twosidefalse
\@mparswitchfalse
\makeatother

\usepackage{graphicx}
\usepackage[utf8]{inputenc}
\usepackage{newtxtext}
\usepackage{newtxmath}
\usepackage{newtxtt}
\usepackage{fontawesome}
\usepackage{courier}
\usepackage{booktabs}
\usepackage{multirow}
\usepackage{tabularx}
\usepackage{makecell}
\usepackage{appendix}
\usepackage{subfigure}

\let\llncssubparagraph\subparagraph
\let\subparagraph\paragraph
\usepackage{titlesec}
\let\subparagraph\llncssubparagraph

\usepackage{scalerel}
\usepackage{tikz}
\usetikzlibrary{svg.path}

\definecolor{orcidlogocol}{HTML}{A6CE39}
\tikzset{
  orcidlogo/.pic={
    \fill[orcidlogocol] svg{M256,128c0,70.7-57.3,128-128,128C57.3,256,0,198.7,0,128C0,57.3,57.3,0,128,0C198.7,0,256,57.3,256,128z};
    \fill[white] svg{M86.3,186.2H70.9V79.1h15.4v48.4V186.2z}
                 svg{M108.9,79.1h41.6c39.6,0,57,28.3,57,53.6c0,27.5-21.5,53.6-56.8,53.6h-41.8V79.1z M124.3,172.4h24.5c34.9,0,42.9-26.5,42.9-39.7c0-21.5-13.7-39.7-43.7-39.7h-23.7V172.4z}
                 svg{M88.7,56.8c0,5.5-4.5,10.1-10.1,10.1c-5.6,0-10.1-4.6-10.1-10.1c0-5.6,4.5-10.1,10.1-10.1C84.2,46.7,88.7,51.3,88.7,56.8z};
  }
}

\newcommand\orcidicon[1]{\href{https://orcid.org/#1}{\mbox{\scalerel*{
\begin{tikzpicture}[yscale=-1,transform shape]
\pic{orcidlogo};
\end{tikzpicture}
}{|}}}}

\usepackage{url}
\usepackage{hyperref} 

\begin{document}
\title{Sequence-to-Sequence Natural Language to Humanoid Robot Sign Language}

\newcommand{\orcidauthorA}{0000-0003-0143-9069} 
\newcommand{\orcidauthorB}{0000-0002-3080-3467} 
\newcommand{\orcidauthorC}{0000-0003-4864-4625} 

\author{Jennifer J. Gago\inst{1}
$^{(\textsc{\faEnvelopeO})}$\orcidID{\orcidicon{0000-0003-0143-9069}}
\and Valentina Vasco\inst{2}\orcidID{\orcidicon{0000-0001-9995-061X}}
\and Bartek \L{}ukawski\inst{1}\orcidID{\orcidicon{0000-0002-1052-3345}}
\and Ugo Pattacini\inst{2}\orcidID{\orcidicon{0000-0001-8754-1632}}
\and \\ Vadim Tikhanoff\inst{2}\orcidID{\orcidicon{0000-0001-5295-3928}}
\and Juan G. Victores\inst{1}\orcidID{\orcidicon{0000-0002-3080-3467}} 
\and Carlos Balaguer\inst{1}\orcidID{\orcidicon{0000-0003-1802-7933}} 
}

\institute{Robotics Lab, Department of Systems Engineering and Automation, University \\ Carlos III of Madrid, Av. Universidad 30, 28911, Leganés, Madrid, Spain\\
\email{jgago@ing.uc3m.es}
\and
iCub Facility, Istituto Italiano di Tecnologia, Via Morego 30, 16163, Genova, Italy
}
\maketitle              
\begin{abstract}
\vspace*{-2.1em} 
This paper presents a study on natural language to sign language translation with human-robot interaction application purposes. By means of the presented methodology, the humanoid robot TEO is expected to represent Spanish sign language automatically by converting text into movements, thanks to the performance of neural networks. Natural language to sign language translation presents several challenges to developers, such as the discordance between the length of input and output data and the use of non-manual markers. Therefore, neural networks and, consequently, sequence-to-sequence models, are selected as a data-driven system to avoid traditional expert system approaches or temporal dependencies limitations that lead to limited or too complex translation systems. To achieve these objectives, it is necessary to find a way to perform human skeleton acquisition in order to collect the signing input data. OpenPose and skeletonRetriever are proposed for this purpose and a 3D sensor specification study is developed to select the best acquisition hardware. 

\end{abstract}
\setcounter{footnote}{0} 

\section{Introduction}

According to the WHO, 466 million people around the world have hearing loss\footnote{Deafness and hearing loss. (2019). Retrieved from \url{https://www.who.int/news-room/fact-sheets/detail/deafness-and-hearing-loss}.}. Within this group, there are 72 million deaf people using 300 different sign languages\footnote{International Day of Sign Languages 23 September. (2019). Retrieved from \url{https://www.un.org/en/events/signlanguagesday/}.}. Today's society is becoming more aware of the importance of the inclusion of everyone and design for all, regardless of people needs. That is why society and science is giving due importance to sign language and around 284 thousand publications can be found to the present, matching sign language and robotics. 

One of the most characteristic aspects about sign language is that the grammatical structure is not the same as in the natural language. That is to say, the order of the words does not coincide with that of the signs when forming sentences. In the case of the Spanish oral language, the order is subject-verb-object; on the other hand, in the Spanish Sign Language (LSE), the order is subject-object-verb. This is the basic order of signs within a sentence, although there are less frequent exceptions. Following this line, the intrinsic difficulty of translations between any natural language and sign language is inherent, not only in the order criteria, but also in the sentences simplification, since there is not a word-sign exact correlation~\cite{spanish}.

Taking this translation complexity into consideration, creating a tool which could be used to translate between natural language and sign language could be a fast and powerful solution to enable communication between signers and non-signers. Beyond this application, the objective of this study is to apply this resource to the humanoid robot TEO, a household companion assistive robot which is able to do the laundry, paint, serve drinks and represent sign language, among other tasks~\cite{gago}. The importance of teaching an assistive robot how to sign lies in the pursuit of a comfortable human robot interaction for all. 

The scientific term for the automated translation from sign language to natural language or vice versa is sign language machine translation, which appeared in 1977, thanks to project RALPH, and involved a finger-spelling robotic hand that could translate English text to American Sign Language finger-spelling \cite{jaffe}. The development of this technology continued with the use of wearable devices such as gloves, which tracked the movements of the hand joints thanks to motion sensors, and it was followed by the use of cameras and the upsurge of computer vision \cite{parton}. The use of cameras represented a considerable progress, since sign languages have unique parameters, frequently known as non-manual markers, which are difficult to track just via motion sensors. Other methods for sign recognition have been used during the past years, such as Hidden Markov Models to analyze the data in a statistical way \cite{starner} \cite{vogler}, and neural networks, which along with computer vision can be used to improve sign recognition accuracy \cite{pigou}. 

Some of the most relevant projects regarding sign language tracking include hardware such as CyberGlobe, a sensored glove that provides up to 22 high-accuracy joint-angle measurements. 
The pattern recognition with the Cyberglobe was initially developed thanks to the Stuttgart Neural Network Simulator in 1999. Even the use of these devices is quite popular, as mentioned before, sign language tracking with wearables is limiting in some cases, since non-manual markers are not easily detected with these devices and, moreover, it may restrain motion. Therefore, a wearable-free solution is pursued in this work. 

The use of avatar software technology is quite widespread in significant projects related to machine translation, such as the Avatar Project\footnote{University, T. (2019). The American Sign Language Avatar Project at DePaul University. Retrieved from \url{http://asl.cs.depaul.edu/}.}, or eSIGN, a commercial solution that produced software tools which allow website and other software developers to augment their applications with signed versions. The eSIGN team states that it is not possible to automate sign language translation, so that projects do not contemplate the possibility of developing an actual translator, but offering punctual translation services\footnote{Introduction to eSIGN. (2019). Retrieved from \url{http://www.visicast.cmp.uea.ac.uk/eSIGN/Introduction.htm\#12}.}. Taking into account the fact that the grammar of sign languages is evidently different from that of spoken languages, the approach to be followed in this case should consider the possibility of exploring alternatives to the traditional translation methodologies. 

Contrary to the proposed sequence-to-sequence method, many of the state of the art projects are rule-based \cite{san} \cite{tokuda}, or statistical-based \cite{san2} \cite{othman}, but it is not easy to find publications in which neural networks are used to make the natural language to sign language translation possible. The main inconvenience in using rule-based models is that they are based in expert systems, so it is not realistic to expect that this system can cover all translation possibilities. The disadvantage of using statistical-based models is that they do not consider temporal dependencies, which are fundamental in sign language representation. The main motivation behind using the proposed sequence-to-sequence method is to advocate data-driven model, or what is the same, let the data find its own place and do not force the data transformation. By using neural networks, it is to be expected that the possibilities of translation increase in relation to the time investment. Furthermore, sequence-to-sequence architecture is appropriate in cases where the length of the input sequence does not match the length of the output data \cite{sutskever}, which is the most frequent case in sign language translations. 

On the other hand, although some papers study the kinematics of the translated signs \cite{mcdonald}, there is a lack of testing with robots, which would be another main contribution presented by the present document, since the output of the current pipeline is robot LSE signing from text in natural language. Consequently, this methodology would present an innovative tool to allow robots to directly communicate in sign language, resulting in a more inclusive human-robot interaction. 

Regarding the means whereby LSE tokens can be transformed to sign language robot execution and pose acquisition, once wearable devices have been discarded, some popular computer vision resources need to be explored and studied. Kinect 3D sensor have been used in some sign language translator project such as the Kinect Sign Language Translator \cite{chai} and other sign language recognition works \cite{zafrulla}. RealSense cameras are also popular in the sign language recognition field, as well as Asus Xtion cameras and Leap Motion devices. Some studies use multiple devices to avoid occlusions or resolution issues. Deep analysis of these devices specifications is required to opt for the best solution, taking into account the specific casuistry of sign language recognition, which requires a precise skeleton tracking of the top body as well as a detailed view of the finger movement. 

Taking into account the previous points, the contribution of this work can be summed up in the three following points:

\begin{itemize}
   \item Direct translation from natural language text to LSE robot execution.
   \item Use of a data-driven translation approach by sequence-to-sequence models.
   \item Testing with a humanoid assistive robot.
\end{itemize}

The remainder of this paper is organised in five sections. Section~II comprises the study of the sequence-to-sequence model. Section~III describes the data acquisition process and how to transform the LSE tokens in robot execution. Section~IV outlines the experiments regarding the sequence-to-sequence model and section~V the ones related to the conversion from LSE tokens to robot execution. Finally, conclusions are drawn in relation to this research.

\section{Sequence-to-Sequence Model} \label{sec:seq2seq}

This first step of the full pipeline accepts natural language text as input. This text is tokenized in a one-hot encoding and fed to a sequence-to-sequence model, which translates it into LSE tokens which are detokenized in Section~\ref{sec3}.

\begin{figure}[h!]
\centering
\includegraphics[width=\textwidth]{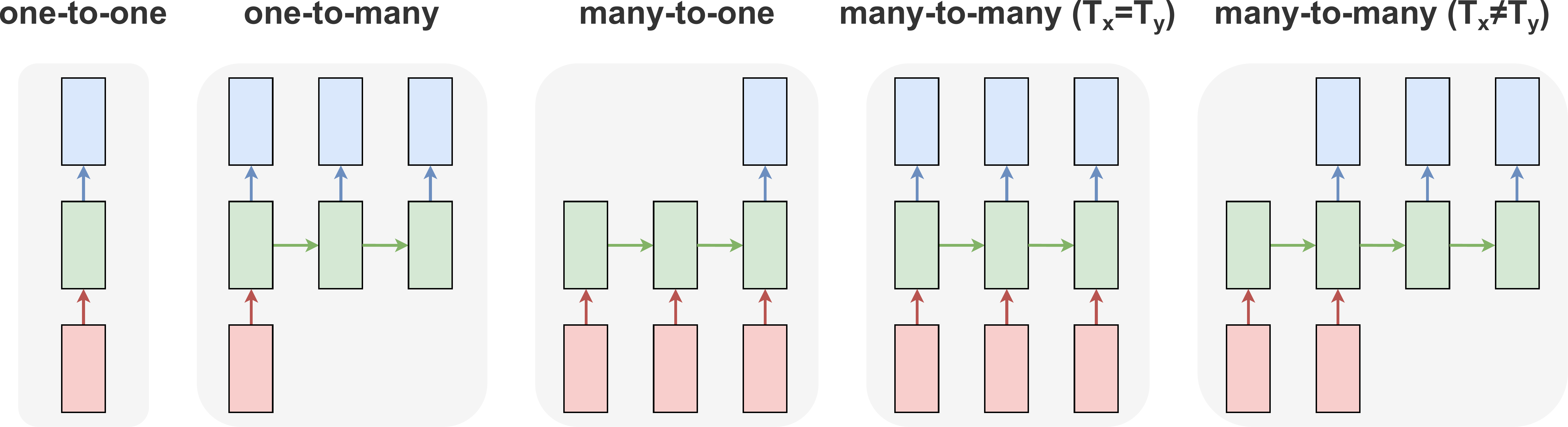}
\caption{Sequence-to-sequence model: several possible layouts.}
\label{fig:rnns} 
\end{figure}

The sequence-to-sequence model is implemented as a many-to-many Recurrent Neural Network (RNN). RNNs are a class of neural network that have the form of a chain of repeating cells, and have been proven to perform extremely well on temporal data,
such as speech recognition, natural language processing, and other areas such as stock trade markets.
The RNN cells are usually laid out horizontally, with time index $t$ advancing from left to right, and inputs on the bottom and outputs on the top. Many different layouts are possible, including stacking several cells up vertically to obtain deeper networks. Depending on the relation between the length of the input sequence ($T_x$) and the length of the output sequence ($T_y$), several different layouts can be 
selected as sequence-to-sequence models, as seen in Figure~\ref{fig:rnns}.


The main unit of a conventional neural network is the perceptron, of input $x$ and output $y = g \big( W x + b \big)$; where $g$ is an activation function (typically Sigmoid, ReLU, ELU, or related), $W$ is the weight matrix, and $b$ is the bias term.
Weight matrices and biases are adjusted during the training phase, and more importantly, they uniquely belong to each corresponding perceptron. However, in RNNs, these weights are typically shared across cells of a given layer. Additionally, the particular layout and connections required for an RNN give rise to alternatives to the conventional perceptron. The most popular alternatives are, namely, the Simple RNN, the Gated Recurrent Unit (GRU), and Long Short-Term Memory (LSTM).

For long sequences, RNNs are prone to long-term dependency problems due to vanishing gradients. The latter mentioned LSTM \cite{hochreiter} is explicitly designed to avoid this problem, by passing a cell state on to the next cell, in addition to the activation. The equations of an LSTM cell are the following:

\begin{equation}
\tilde{c}^{<t>} = \tanh \big( W_c [a^{<t-1>}, x^{<t>}] + b_c \big)
\end{equation}
\begin{equation}
\Gamma_u^{<t>} = \sigma \big( W_u [a^{<t-1>}, x^{<t>}] + b_u \big)
\end{equation}
\begin{equation}
\Gamma_f^{<t>} = \sigma \big( W_f [a^{<t-1>}, x^{<t>}] + b_f \big)
\end{equation}
\begin{equation}
\Gamma_o^{<t>} = \sigma \big( W_o [a^{<t-1>}, x^{<t>}] + b_o \big)
\end{equation}
\begin{equation}
c^{<t>} = \Gamma_u^{<t>} \ast \tilde{c}^{<t>} + \Gamma_f^{<t>} \ast c^{<t-1>}
\end{equation}
\begin{equation}
a^{<t>} = \Gamma_o^{<t>} \ast c^{<t>}
\end{equation}

Where $c$ is the cell state, and $\tilde{c}$ is the cell state that is ``candidate'' for the next update.
$W_c$, $W_u$, $W_f$, and $W_o$ are the state, update, forget, and output weight matrices, respectively, and $b_c$, $b_u$, $b_f$, and $b_o$ are the bias terms.
$\Gamma_u$, $\Gamma_f$, $\Gamma_o$ are the update, forget and output gates, respectively.
$a$ is the hidden activation output. Finally, ``$\ast$'' refers to the Hadamard element-wise product.

\begin{figure}[h!]
\centering
\includegraphics[trim={1em, 1em, 1em, 1em},width=0.8\textwidth]{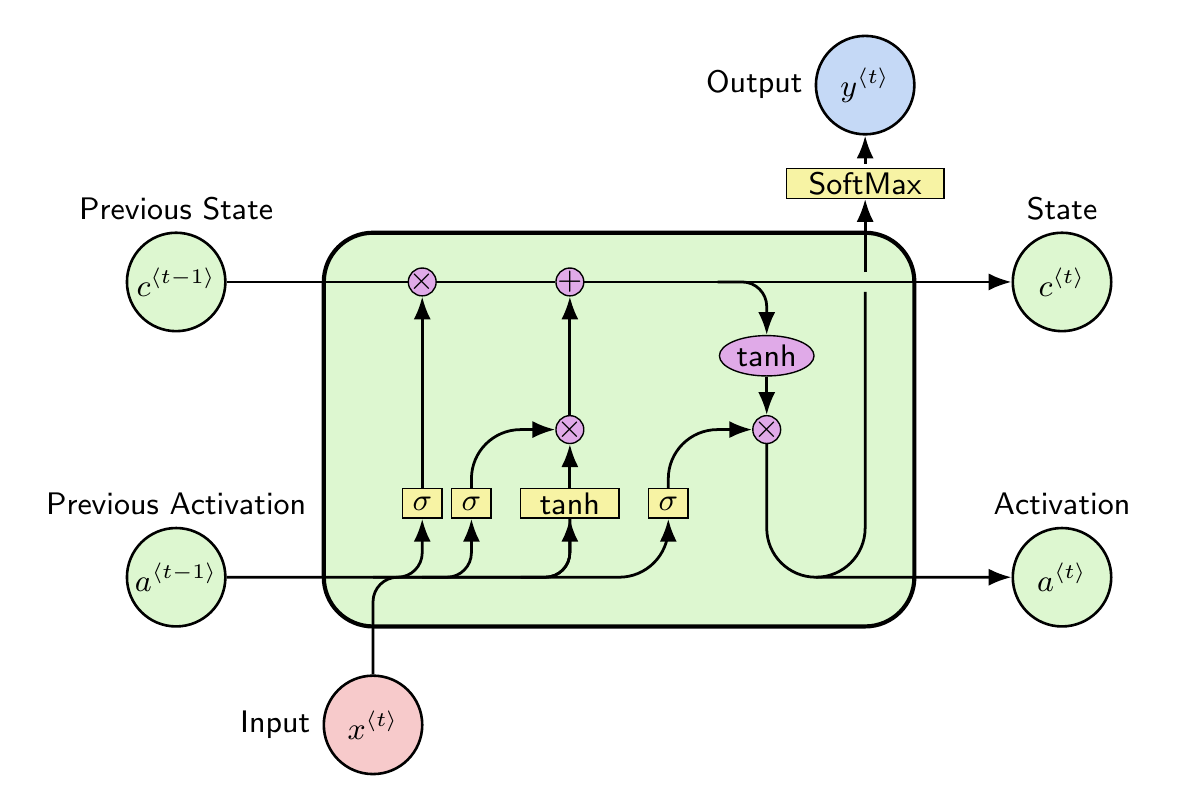}
\caption{The LSTM cell internals.}
\label{fig:lstm}
\end{figure}

A depiction of these LSTM internals can be seen in Figure~\ref{fig:lstm}, which additionally includes a fully connected layer with a SoftMax activation that is usually stacked to obtain the cell output when in absence of a more complex layout.

These networks are typically trained in a supervised fashion, providing paired input ($x$) and output ($y$) examples.
The output predictions ($\tilde{y}$) are computed via a Forward Propagation process, whereas the weights and bias updates are computed during a Back Propagation Through Time process. Optimization algorithms that are used for conventional neural networks, such as Stochastic Gradient Descent or Adam, may be used for this process.

\section{From LSE Tokens to Robot Sign Execution}
\label{sec3}

This second step of the full pipeline accepts LSE tokens, which are 
converted into robot sign execution via a look-up-table (LUT). The robot movements are executed sequentially, in order of arrival from the sequence-to-sequence output of Section~\ref{sec:seq2seq}.

The LUT is populated in an offline fashion, recording the joint space configuration (or sequence of configurations) that corresponds to each individual LSE token.
The joint space configuration is obtained from human demonstrations via a two-fold implementation: a 2D skeleton acquisition system explained in Section \ref{sec:openpose}, which is then combined with depth data to obtain 3D skeleton poses as explained in Section \ref{sec:skeletonretriever}. The corresponding joint space configurations are obtained via conventional inverse kinematics.


\subsection{2D Skeleton Acquisition: \emph{OpenPose}} \label{sec:openpose}

\begin{figure}[h!]
\centering
\includegraphics[width=0.7\textwidth]{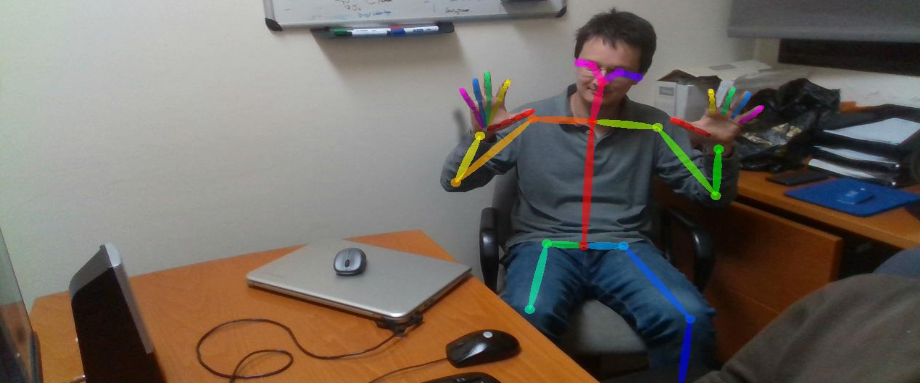}
\caption{OpenPose 2D skeleton pose estimation.}
\label{fig:openpose}
\end{figure}

OpenPose is an open source, cross-platform, real-time library for multi-person 2D pose estimation including body, foot, hand and facial keypoints \cite{cao}, Figure~\ref{fig:openpose}.
It follows a bottom-down approach that outperforms similar 2D body pose estimation libraries (e.g. Mask R-CNN, Alpha-Pose) and provides a reasonable trade-off between speed and accuracy.
Several challenges are tackled such as: spatial interference between people and parts due to occlusions and contact; unknown number, location and scale of people on the image; run-time complexity increasing with the number of people.

This library achieves high accuracy and performance regardless of the number of people in the image by using a non-parametric representation of 2D vector fields that encode the position and orientation of body parts over the image domain and their degree of association, referred to as Part Affinity Fields (PAFs), in order to learn to relate them to each individual.

OpenPose implements a convolutional neural network (CNN) initialized by the 10 first layers of VGG-19 and fine-tuned.
The iterative architecture performs intermediate supervision at each stage and jointly learns part detection and association.
First, part-to-part association is encoded by predicting affinity fields with refinement at each subsequent stage.
Then, confidence maps are predicted for body part detection, starting on top of the latest and most refined PAF predictions.

The overall pipeline processes an input color image to obtain 2D anatomical keypoints for each person found.
A set of confidence maps and a set of vector fields of PAFs are predicted by the feedforward network and consecutively parsed by greedy inference.
The PAFs help assemble detected parts by preserving both position and orientation, thus eliminating false associations.

OpenPose can run on multiple operating systems and different hardware architectures (CUDA GPUs, OpenCL GPUs, CPU-only).
It provides tools for acquisition, visualization and output file generation.
Other vendors require that the user implements its own pipeline and do not offer a combined keypoint detector for body and other parts (hands, face).

\subsection{3D Skeleton Acquisition: \emph{skeletonRetriever}} \label{sec:skeletonretriever}

\begin{figure}[h!]
\centering
\includegraphics[width=0.9\textwidth]{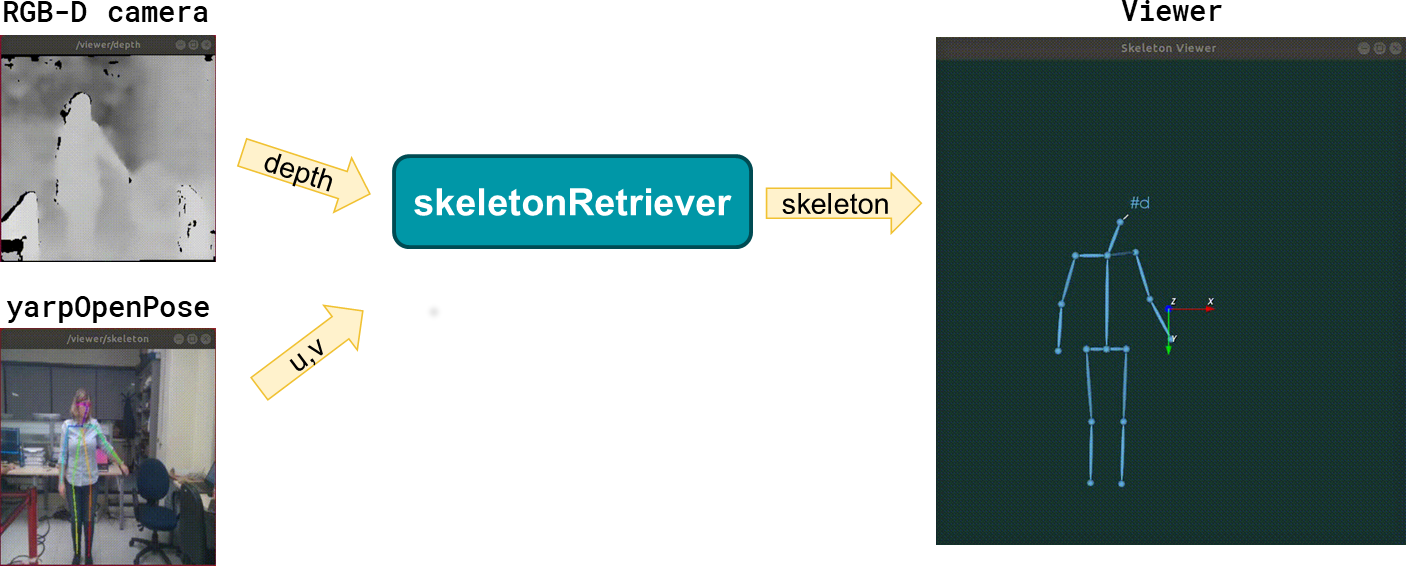}
\caption{Pipeline for 3D skeleton acquisition.
}
\label{fig:fig1-pipeline}
\end{figure}

The 3D representations of human skeletons are generated by combining the 2D locations of the keypoints provided by the skeleton detector based on \emph{OpenPose} with the depth retrieved from the camera. The acquisition pipeline\footnote{The code is freely available on GitHub at \url{https://github.com/robotology/assistive-rehab}.} is shown in Figure~\ref{fig:fig1-pipeline} and consists of a set of modules interconnected together on a YARP network.

The 3D reconstruction is performed in such a way that we can guarantee with respect to the task at hand a certain level of robustness to keypoints self-occlusions and mismatches occurring between detected keypoints and noisy depth contours.

\begin{figure}[h!]
\centering
\includegraphics[width=1\textwidth]{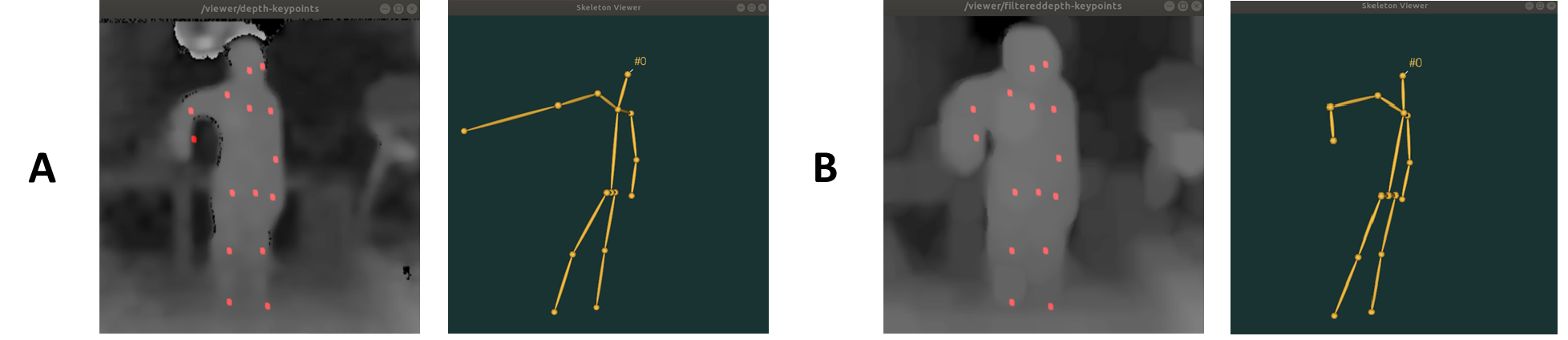}
\caption{Depth filtering. 
}
\label{fig:fig2-depth-filtering}
\end{figure}

In detail, dilation is applied to depth images in the first stage of the algorithm to obtain corresponding frames where the probability that a keypoint is detected in the close neighbourhood of a depth contour representing a human body part is significantly reduced. The noise present in the depth likely causes the 2D keypoints that are detected very close to that contour to be projected in 3D estimates that are far from their actual positions. To circumvent this phenomenon, it suffices to dilate depth contours of the regions closer to the camera with the aim to increase the distance margin between the keypoints and the nearest border. This process is illustrated in Figure~\ref{fig:fig2-depth-filtering}: as a result of dilation, the keypoint representing the hand and lying close to the contours of the arm in the original depth image (see Figure~\ref{fig:fig2-depth-filtering}-A) is ensured to stay consistently within the new filtered region (see Figure~\ref{fig:fig2-depth-filtering}-B). Additionally, dilation does contribute to fill in holes that may affect the input depth.

Thereby, the estimate of the 3D keypoint $p_i$ is readily determined by resorting to the projection of the $i$-th 2D keypoint coordinates $(u,v)_i$ identified by the detector and the relative filtered depth $d_i$, using the classical pinhole camera model characterized by the focal $f$ and the image width and height $(w,h)$:

\begin{equation}
p_i=d_i*{\begin{pmatrix} \frac{u_i-w⁄2}{f} \\ \frac{v_i-h⁄2}{f}  \\ 1\end{pmatrix}}
\end{equation}

\begin{figure}[h!]
\centering
\includegraphics[width=0.9\textwidth]{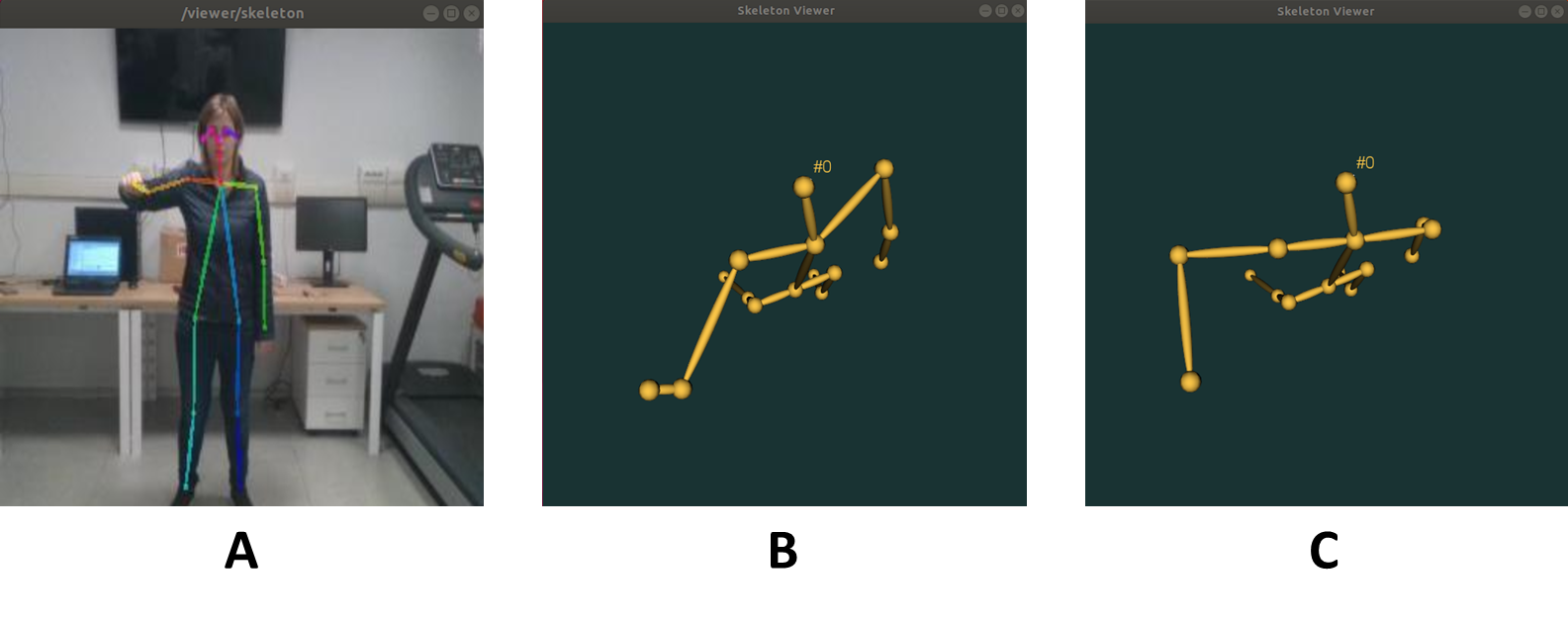}
\caption{Optimization applied to skeleton limbs. 
}
\label{fig:fig3-optimization}
\end{figure}

The resulting 3D estimates undergo a further smoothing that takes place in the form of a median filtering of a specified order (typically 4÷5), which is responsible for providing a higher grade of space-temporal contiguity among the keypoints representing the same parts of the skeleton as extracted from subsequent frames. 
Finally, we also implement an optimization technique to cope with keypoints self-occlusions as depicted in Figure~\ref{fig:fig3-optimization}. In case two or more keypoints are aligned with the viewpoint of the camera, their corresponding depth values get overlapped thus impeding a correct 3D projection of the associated body parts of the skeleton. To find better estimates for the occluded depth values, we solve an optimization problem where the latent depths are adjusted in order to minimize the quadratic difference between the reconstructed lengths of the skeleton limbs and those same lengths observed during an initial calibration, when we assume that the limbs are fully visible (i.e. no overlap among keypoints). We run multiple minimizations to address individual skeleton limbs (i.e. arms and/or legs) having overlapped keypoints. This procedure is carried out in real-time using Ipopt \cite{watcher}, which is a state-of-the-art library for nonlinear constrained optimization. 

Although this approach is local and lightweight compared with other works that employ global optimization (e.g. VNect \cite{mehta}), it allows to reconstruct skeletons that are robust against impairments involving noisy depth and keypoints self-occlusions, as demonstrated by the successful deployment of the 3D acquisition pipeline within the whole framework.

\section{Experiments: Sequence-to-Sequence Model}

The sequence-to-sequence model accepts natural language text sequences as input, and outputs translated LSE token sequences.
To achieve this task, both the input and output sequences are tokenized at word-level, reserving additional tokens for exclamation and question marks, and each token is encoded via a one-hot vector.

\begin{figure}[h!]
\centering
\subfigure[Learning rate $\alpha = 0.0001$]{\label{fig:seq2seq-loss-a}\includegraphics[trim={5em, 1.5em, 3.5em, 1.5em},width=0.47\textwidth]{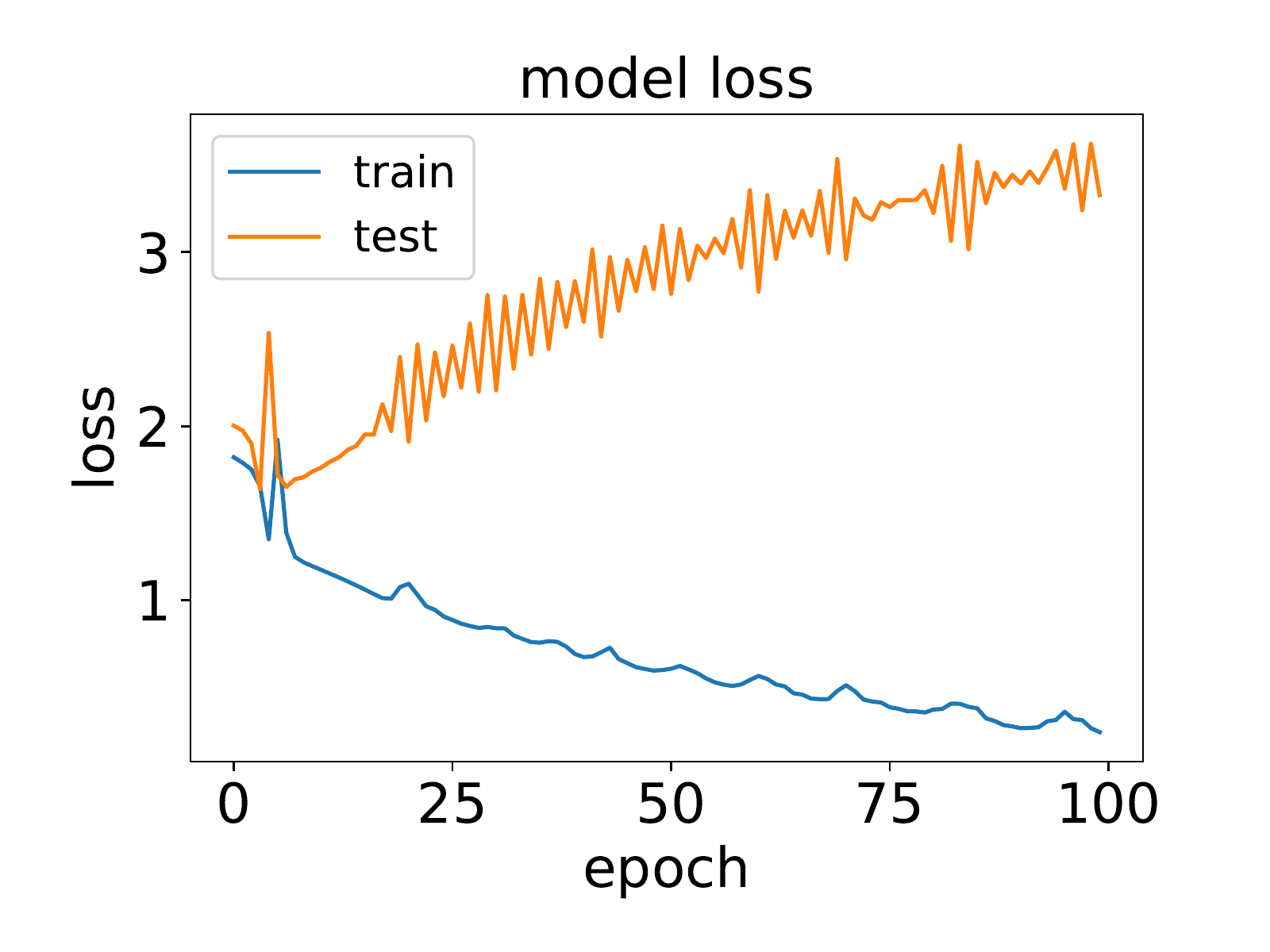}}
\quad
\subfigure[Learning rate $\alpha = 0.001$]{\label{fig:seq2seq-loss-b}\includegraphics[trim={3.5em, 1.5em, 5em, 1.5em},width=0.47\textwidth]{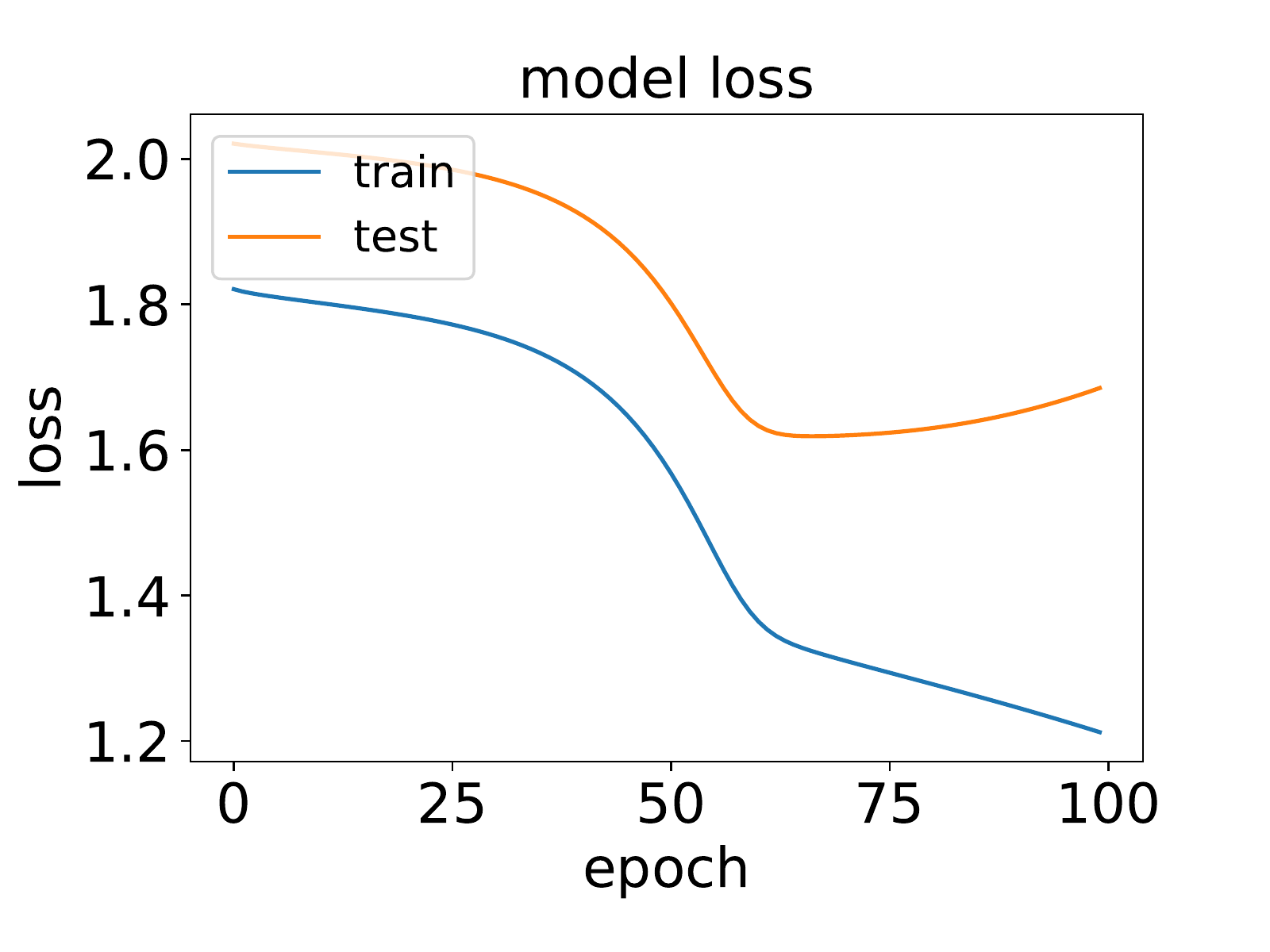}}
\caption{Training loss and cross-validation loss during training.}
\label{fig:seq2seq-loss}
\end{figure}

The model corresponds to a $T_x \neq T_y$ many-to-many layout, see Figure \ref{fig:rnns}, which enables input and output sequences of arbitrary lengths. A single layer of LSTM cells with 256 hidden activation is used, which are output as predictions through a fully-connected layer with a SoftMax activation.
A categorical cross-entropy loss is used for an RMSprop optimizer with a $\beta=0.9$. 
Training data is depicted in Appendix~\ref{sec:app-training-data}.
The model is trained for 100 epochs.
For cross-validation, a 20\% split is used.
%
%
Training and cross-validation loss during training are depicted in Figure~\ref{fig:seq2seq-loss}.
Prediction results with learning rate $\alpha = 0.0001$ are listed in Table~\ref{tab:test}.


\begin{table}[h!]
\centering
\caption{Predicted output for the first 10 examples with learning rate $\alpha = 0.0001$.}
\label{tab:test}
\begin{tabularx}{\textwidth}{XX}
\toprule
\textbf{Natural Language Text Input ($x$)} & \textbf{LSE Token Output Prediction ($\tilde{y}$)} \\
\midrule
¿Qué tal? & Tú Bien \\
Estoy bien pero tengo sueño & Bien Dormir  \\
¿Tú vas al colegio? & Colegio \\
Venga, levántate, que tienes que ir al colegio & Venga Levantar Tú Colegio Ir \\
¿Qué has hecho en el colegio? & Colegio \\
¿Qué deberes tienes que hacer? & Colegio \\
¿Cuántos años tienes? & Edad Cuántos \\
Ana, ¿quién es Ana? & Ana Quién \\
¿Cómo te llamas? & Tú Nombre Cuál \\
Me llamo… & Mi Nombre \\
\bottomrule
\end{tabularx}
\end{table}

\section{Experiments: From LSE Tokens to Robot Sign Execution}

\begin{figure}[h!]
\centering
\includegraphics[width=0.75\textwidth]{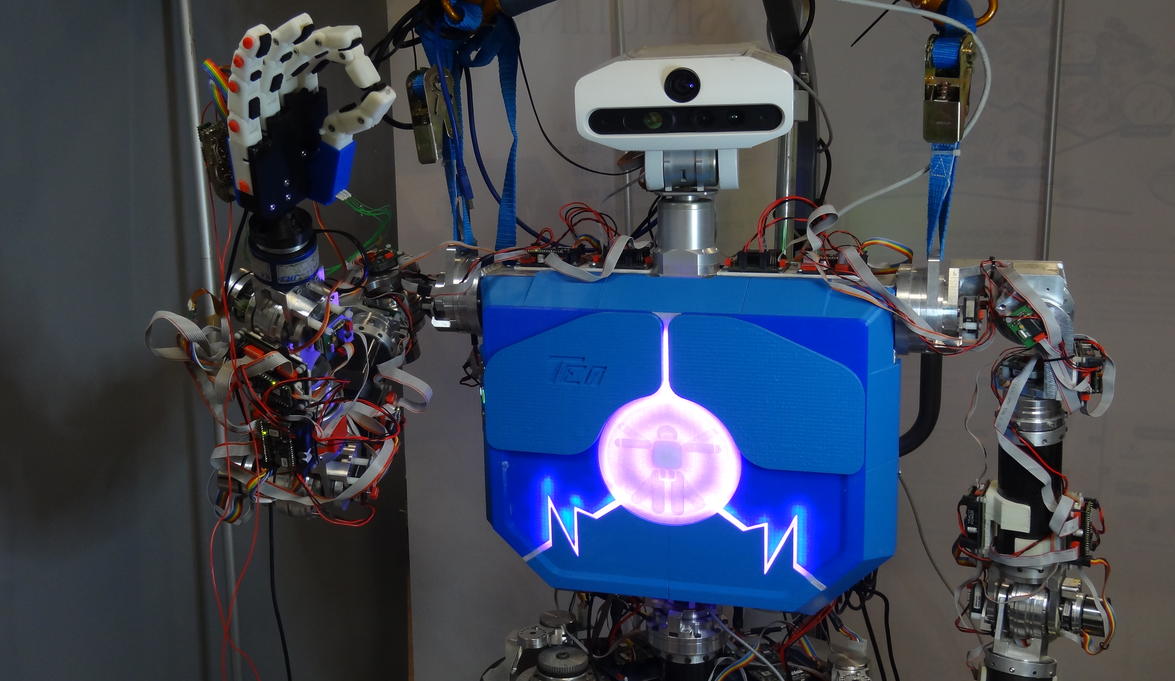}
\caption{TEO robot signing via LUT from LSE tokens to execution.}
\label{fig:teo}
\end{figure}

The LUT accepts translated LSE token sequences as input and performs robot signing on the TEO humanoid robot, Figure~\ref{fig:teo}.
Regarding skeleton acquisition for populating the LUT, relevant specifications of selected RGB-D sensors are gathered in Table~\ref{tab:sensors}.

\begin{table}[h!]
\centering
\caption{RGB-D sensor device specifications.}
\label{tab:sensors}
\begin{tabular}{@{}lcccccc@{}}
\toprule
\multirow{2}{*}[-0.25em]{Device} & \multicolumn{2}{c}{Camera Resolution} & \multirow{2}{*}{\thead{Depth \\ Accuracy}} & \multirow{2}{*}{\thead{Depth \\ Distance}} & \multirow{2}{*}{\thead{Field of View \\ (H x V)}} \\ \cmidrule(l){2-3}
& RGB (30 fps) & Depth & & & \\ \midrule
RealSense SR300 & $1920 \times 1080$ & $640 \times 480$ & 1 mm & 0.2--1.5 m & $73^{\circ}\times59^{\circ}$ \\
RealSense D435 & $1920 \times 1080$ & $1280 \times 720$ & 2 mm & 0.2--10 m & $69.4^{\circ} \times 42.5^{\circ}$ \\
Kinect 1 & $640 \times 480$ & $320 \times 240$ & \textemdash & 0.4--4.5 m & $57^{\circ} \times 43^{\circ}$ \\
Kinect 2 & $1920 \times 1080$ & $512 \times 424$ & \textemdash & 0.5--4.5 m & $70^{\circ} \times 60^{\circ}$ \\
Asus Xtion Pro Live & $1280 \times 1024$ & $640 \times 480$ & \textasciitilde{}10 mm & 0.8--3.5 m & $58^{\circ} \times 47^{\circ}$ \\
Leap Motion & \textemdash & $1600 \times 1440$ & 0.01 mm & 0.025--0.6 m & $135^{\circ}$ average \\ \bottomrule
\end{tabular}
\end{table}

The RealSense D435 sensor was selected as it combines high resolution and depth accuracy along with a reasonable depth distance. 
The Kinect devices were discontinued and therefore not considered. 

\section{Conclusions}

This study shows the development procedure of a natural language text to sign language translator via neural networks, tested on the humanoid robot TEO. The results obtained by applying the sequence-to-sequence model, which are LSE token sequences, have shown excellent results, predicting most terms correctly.

The use of the presented sequence-to-sequence model is justified by the proof-of-concept point of view.
Future prospects may focus on attention mechanisms and augmented RNNs, that will allow neural networks to work with larger sequences of data including text and video, which is especially useful for this topic, since sign language is basically a visual language \cite{google}. These architectures will verify that the proposed translator concept is susceptible of being exploited in a useful way, confirming the robustness of the system.

\newpage

\appendix

\titleformat{\section}{\normalfont\large\bfseries}{Appendix\ \Alph{section}.\quad}{0pt}{\large}

\section{Training Data for Supervised Learning} \label{sec:app-training-data}

\begin{table}[h!]
\centering
\begin{tabularx}{\textwidth}{XX}
\toprule
\textbf{Training Input ($x$)} & \textbf{Training Output ($y$)} \\
\midrule
¿Qué tal? & Tú Bien \\
Estoy bien pero tengo sueño & Bien Dormir \\
¿Tú vas al colegio? & Tú Colegio Ir \\
Venga, levántate, que tienes que ir al colegio & Venga Levantar Tú Colegio Ir \\
¿Qué has hecho en el colegio? & Colegio Hacer Qué \\
¿Qué deberes tienes que hacer? & Colegio Ejercicios Qué \\
¿Cuántos años tienes? & Edad Cuántos \\
Ana, ¿quién es Ana? & Ana Quién \\
¿Cómo te llamas? & Tú Nombre Cuál \\
Me llamo… & Mi Nombre \\
Buenos días & Bien Día \\
Buenas tardes & Bien Tarde \\
Buenas noches & Bien Noche \\
Yo tengo X años & Yo Edad X \\
¡Vamos a comer! & Vamos Comer \\
¡A despertarse! & Vamos Despertarse \\
¡Vamos a dormir! & Vamos Dormir \\
¡Vamos a la ducha! & Vamos Ducha \\
¡Vamos a trabajar! & Vamos Trabajar \\
Estoy muy enfadado & Yo Enfadado \\
Te quiero mucho & Yo Amor \\
Dame un abrazo & Vamos Amor \\
¡Vamos a estudiar! & Vamos Estudiar \\
¡Vamos a hacer los deberes! & Vamos Deberes \\
¡Vamos a la calle! & Vamos Calle \\
¡Vamos a pasear! & Vamos Pasear \\
¿Estas cansado? & Tú Cansado \\
¿Estas malo? & Tú Enfermo \\
¿Estás mejor? & Tú Mejor \\
¿Me ayudas? & Ayúdame \\
¿Qué quieres comer? & Tú Comer Qué \\
¿Qué te apetece hacer? & Tú Hacer Qué \\
¿Qué te duele? & Tú Doler Qué \\
¿Tienes hambre? & Tú Hambre \\
¿Tienes miedo? & Tú Miedo \\
¡Cuidado! Eso puede romperse! & Cuidado Eso Romper \\
Dame eso, por favor & Dame Por favor Dame \\
Dame la mano & Dame mano \\
Dame un beso & Tú Beso \\
\bottomrule
\end{tabularx}
\end{table}


\begin{thebibliography}{8}

\bibitem{spanish}
Herrero Blanco, Á., Abellán, A., \& José, J. (1999). Fonología y escritura de la lengua de signos española. ELUA. Estudios de Lingüística, N. 13 (1999); pp. 89--116.

\bibitem{gago}
Gago, J., Victores, J., \& Balaguer, C. (2019). Sign Language Representation by TEO Humanoid Robot: End-User Interest, Comprehension and Satisfaction. Electronics, 8(1), 57.

\bibitem{jaffe}
Jaffe, D. L. (1994). Evolution of mechanical fingerspelling hands for people who are deaf-blind. Journal of rehabilitation research and development, 31(3), 236--244.

\bibitem{parton}
Parton, B. S. (2005). Sign language recognition and translation: A multidisciplined approach from the field of artificial intelligence. Journal of deaf studies and deaf education, 11(1), 94--101.

\bibitem{starner}
Starner, T., \& Pentland, A. (1997). Real-time american sign language recognition from video using hidden markov models. In Motion-Based Recognition (pp. 227--243). Springer, Dordrecht.

\bibitem{vogler}
Vogler, C., \& Metaxas, D. (1999). Parallel hidden markov models for american sign language recognition. In Proceedings of the Seventh IEEE International Conference on Computer Vision (Vol. 1, pp. 116--122). IEEE.

\bibitem{pigou}
Pigou, L., Dieleman, S., Kindermans, P. J., et al. (2014, September). Sign language recognition using convolutional neural networks. In European Conference on Computer Vision (pp. 572--578). Springer, Cham.

\bibitem{san}
San-Segundo, R., Barra, R., Córdoba, R., et al. (2008). Speech to sign language translation system for Spanish. Speech Communication, 50(11-12), 1009-1020.

\bibitem{tokuda}
Tokuda, M., \& Okumura, M. (1998). Towards automatic translation from japanese into japanese sign language. In Assistive Technology and Artificial Intelligence (pp. 97-108). Springer, Berlin, Heidelberg.

\bibitem{san2}
San-Segundo, R., Montero, J. M., Córdoba, R., et al. (2012). Design, development and field evaluation of a Spanish into sign language translation system. Pattern Analysis and Applications, 15(2), 203-224.

\bibitem{othman}
Othman, A., \& Jemni, M. (2011). Statistical sign language machine translation: from English written text to American sign language gloss. arXiv preprint arXiv:1112.0168.

\bibitem{sutskever}
Sutskever, I., Vinyals, O., \& Le, Q. V. (2014). Sequence to sequence learning with neural networks. In Advances in neural information processing systems (pp. 3104-3112).

\bibitem{mcdonald}
McDonald, J., Wolfe, R., Schnepp, J., et al. (2016). An automated technique for real-time production of lifelike animations of American Sign Language. Universal Access in the Information Society, 15(4), 551-566.

\bibitem{chai}
Chai, X., Li, G., Lin, Y., et al. (2013, April). Sign language recognition and translation with kinect. In IEEE Conf. on AFGR (Vol. 655).

\bibitem{zafrulla}
Zafrulla, Z., Brashear, H., Starner, T., et al. (2011, November). American sign language recognition with the kinect. In Proceedings of the 13th international conference on multimodal interfaces (pp. 279-286). ACM.

\bibitem{hochreiter}
Hochreiter, S., \& Schmidhuber, J. (1997). Long Short-Term Memory. Neural Computation. 9 (8): 1735--1780.


\bibitem{cao} 
Cao, Z., Hidalgo, G., Simon, T., et al. (2018). OpenPose: realtime multi-person 2D pose estimation using Part Affinity Fields. arXiv preprint arXiv:1812.08008.

\bibitem{watcher}
Wätcher, A., Biegler, L.T. “On the Implementation of a Primal-Dual Interior Point Filter Line Search Algorithm for Large-Scale Nonlinear Programming”, Mathematical Programming 106 (1): pp. 25--57, 2006.

\bibitem{mehta}
Mehta, D., et al. “VNect: Real-time 3D Human Pose Estimation with a Single RGB Camera”, ACM Transactions on Graphics 36 (4), 2017.

\bibitem{google}
Britz, D., Goldie, A., Luong, M. T., \& Le, Q. (2017). Massive exploration of neural machine translation architectures. arXiv preprint arXiv:1703.03906.



\end{thebibliography}
\end{document}